\documentclass[pdflatex,sn-mathphys-num]{sn-jnl}% Math and Physical Sciences Numbered Reference Style 
\usepackage{graphicx}%
\usepackage{multirow}%
\usepackage{amsmath,amssymb,amsfonts}%
\usepackage{amsthm}%
\usepackage{mathrsfs}%
\usepackage[title]{appendix}%
\usepackage{xcolor}%
\usepackage{textcomp}%
\usepackage{manyfoot}%
\usepackage{booktabs}%
\usepackage{algorithm}%
\usepackage{algorithmicx}%
\usepackage{algpseudocode}%
\usepackage{listings}%
\theoremstyle{thmstyleone}%
\theoremstyle{thmstyletwo}%

\theoremstyle{thmstylethree}%

\begin{document}

\title[Article Title]{On The Uncertainty Principle of Neural Networks}

%%=============================================================%%
%% GivenName	-> \fnm{Joergen W.}
%% Particle	-> \spfx{van der} -> surname prefix
%% FamilyName	-> \sur{Ploeg}
%% Suffix	-> \sfx{IV}
%% \author*[1,2]{\fnm{Joergen W.} \spfx{van der} \sur{Ploeg} 
%%  \sfx{IV}}\email{iauthor@gmail.com}
%%=============================================================%%

\author*[1]{\fnm{Jun-Jie} \sur{Zhang}}\email{zjacob@mail.ustc.edu.cn}\equalcont{These authors contributed equally to this work.}

\author[1]{\fnm{Dong-Xiao} \sur{Zhang}}\equalcont{These authors contributed equally to this work.}

\author[1]{\fnm{Jian-Nan} \sur{Chen}}

\author[2]{\fnm{Long-Gang} \sur{Pang}}

\author*[3]{\fnm{Deyu} \sur{Meng}}\email{dymeng@mail.xjtu.edu.cn}

\affil*[1]{\orgdiv{Division of Computational physics and Intelligent modeling}, \orgname{Institute of Nuclear Technology}, \orgaddress{\city{Xi'an}, \postcode{710024}, \state{Shaanxi}, \country{China}}}

\affil[2]{\orgdiv{Key Laboratory of Quark \& Lepton Physics of Ministry of Education}, \orgname{Central China Normal University}, \orgaddress{\city{Wuhan}, \postcode{30079}, \state{Hubei}, \country{China}}}

\affil*[3]{\orgdiv{School of Mathematics and Statistics and Ministry of Education Key Lab of Intelligent Networks and Network Security}, \orgname{Xi'an Jiaotong University}, \orgaddress{\city{Xi'an}, \postcode{710049}, \state{Shaanxi}, \country{China}}}

%%==================================%%
%% Sample for unstructured abstract %%
%%==================================%%

\abstract{In this study, we explore the inherent trade-off between accuracy and robustness in neural networks, drawing an analogy to the uncertainty principle in quantum mechanics. We propose that neural networks are subject to an uncertainty relation, which manifests as a fundamental limitation in their ability to simultaneously achieve high accuracy and robustness against adversarial attacks. Through mathematical proofs and empirical evidence, we demonstrate that this trade-off is a natural consequence of the sharp boundaries formed between different class concepts during training. Our findings reveal that the complementarity principle, a cornerstone of quantum physics, applies to neural networks, imposing fundamental limits on their capabilities in simultaneous learning of conjugate features. Meanwhile, our work suggests that achieving human-level intelligence through a single network architecture or massive datasets alone may be inherently limited. Our work provides new insights into the theoretical foundations of neural network vulnerability and opens up avenues for designing more robust neural network architectures.}

\keywords{Accuracy-robustness trade-off, Machine Learning Vulnerability, Theoretical Foundations, Interdisciplinary Physics, Artificial Intelligence}

%%\pacs[JEL Classification]{D8, H51}

%%\pacs[MSC Classification]{35A01, 65L10, 65L12, 65L20, 65L70}

\maketitle

\section{Introduction}

\begin{figure}
\begin{centering}
\includegraphics[scale=0.44]{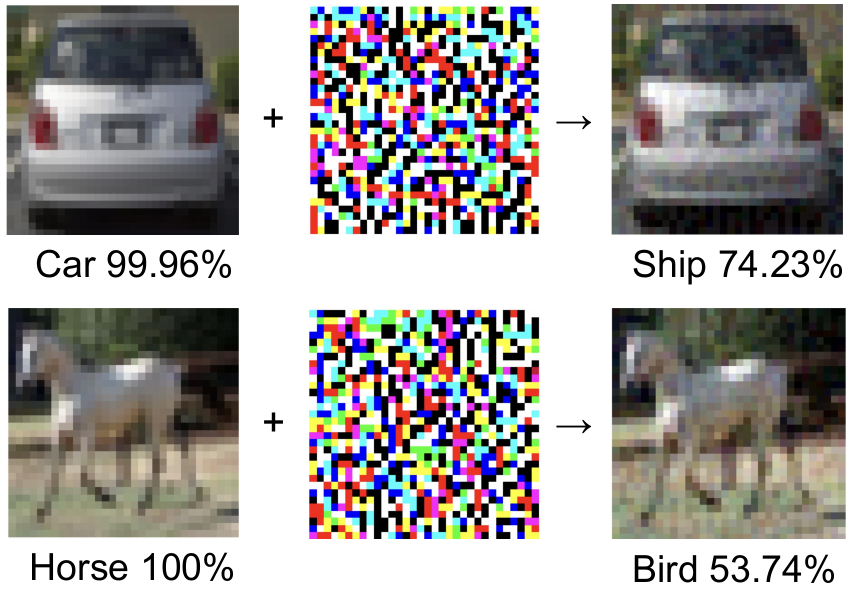}
\par\end{centering}
\caption{Adding an imperceptible non-random noise to the images, the network will fail to predict the correct label.\label{fig:FGSM-attack-reduces} The trained network on Cifar-10
gives a 89.37\% accuracy on the test set and only 9.39\% accuracy on the slightly attacked images (category confidence is labeled below each image). The attack, called FGSM (Fast Gradient Sign Method) \cite{goodfellow2015explaining}, is achieved via the transformation
$X = X_{0}+\epsilon\cdot\text{sign}(\nabla_{X}l(f(X,\theta),Y^{*})| _{X=X_0})$,
where $X_{\text{0}}$ denotes the input images of the training set and and $Y^{*}$
is the true label for image $X_{0}$. $X$ denotes the image to be attacked,
$\epsilon=8/255$ and $\text{sign}(\nabla_{X}l(f(X,\theta),Y^{*})| _{X=X_0})$
gives the non-random noise with $l(f(X,\theta),Y^{*})$
denoting the loss function. $\theta$ denotes the weights of the trained network.}
\end{figure}

Artificial intelligence (AI), particularly deep neural networks, has revolutionized a wide range of applications, for instance, video generation \cite{wang2023videogen}, language models \cite{brown2020language}, protein folding \cite{abramson2024alphafold}, quantum system analysis \cite{RevModPhys.91.045002}, healthcare \cite{topol2019healthcare}, and autonomous driving \cite{bojarski2016tesla}. These advancements highlight the transformative potential of AI across diverse domains, pushing the boundaries of what machines can achieve in both scientific research and real-world applications.

However, despite their remarkable success, a growing body of research has revealed a critical vulnerability: well-designed and trained AI models exhibit significant fragility when confronted with subtle, non-random perturbations \cite{zhang2024airobust, doi:10.34133/icomputing.0088, 759851e20d2e47aaad2a560211f6a126,8578273,jia-liang-2017-adversarial,chen2018attacking,carlini2018audio,xu2012sparse,pmlr-v148-benz21a,morcos2018on}. These perturbations, known as adversarial attacks (Fig. \ref{fig:FGSM-attack-reduces}), involve adding imperceptible noise to input data, which can drastically degrade the performance of a neural network.

Compared to classical models, the vulnerability of deep neural networks exhibits several distinctive characteristics: the impact of adversarial perturbations is both unpredictable and profound, often leading to catastrophic failures in model performance \cite{759851e20d2e47aaad2a560211f6a126,8578273}; moreover, higher-accuracy networks tend to be more susceptible to such attacks \cite{jia-liang-2017-adversarial,chen2018attacking}; efforts to enhance robustness, such as adversarial training where perturbed data is incorporated into the training set \cite{NEURIPS2020_61d77652,arani2020adversarial,arcaini2021roby}, often result in a trade-off, with improved robustness coming at the cost of reduced accuracy. These interconnected observations suggest an inherent tension between accuracy and robustness in neural networks, indicating a fundamental limitation rooted in these models.

Classical approximation theorems assert that neural networks can approximate continuous functions with arbitrary precision \cite{cybendo1992approximations,HORNIK1989359,10.1162/neco.1989.1.4.502,737488}, implying that stable functions should, in theory, yield stable solutions. Yet, the persistent accuracy-robustness trade-off challenges this assumption. If this trade-off arises solely from network architecture or data acquisition, it might be mitigated through improved engineering. However, if it is intrinsic to the foundational principles of deep learning, a deeper theoretical understanding is required. To date, no definitive explanation for this phenomenon has been established.

In this study, we propose a theoretical framework that attributes the accuracy-robustness trade-off to an uncertainty principle analogous to that in quantum mechanics, which posits that certain pairs of properties cannot be simultaneously determined with arbitrary precision. Translating this concept to neural networks, we argue that a network cannot simultaneously extract two complementary features with maximal accuracy. We substantiate this framework with both analytical and experimental evidence, demonstrating that this uncertainty principle is a universal characteristic of neural networks. 

Our findings reveal that the complementarity principle \cite{Bohr1950}, a cornerstone of quantum physics, applies to neural networks, imposing fundamental limits on their capabilities in simultaneous learning of conjugate features. Meanwhile, our work suggests that achieving human-level intelligence through a single network architecture or massive datasets alone may be inherently limited. 

\section{Observation: Neural networks are vulnerable under small imperceptible attacks}\label{sec2}

As illustrated in Fig. \ref{fig:FGSM-attack-reduces}, trained neural networks can fail under non-random attacks. These phenomena, typically constructed using the gradient of the loss function with respect to the inputs, differ significantly from classical algorithms, which are robust to small input variations. Furthermore, the vulnerability of these neural networks is related to their accuracy — more accurate neural networks tend to be more sensitive to gradient-based attacks, indicating a trade-off between accuracy and robustness \cite{10.1007/978-3-030-01258-8_39,xu2012sparse,arcaini2021roby,pmlr-v97-zhang19p,goodfellow2015explaining,tsipras2019robustness,antun2021can}.

\subsection{\textbf{Phenomena in image classification}}

\begin{figure}
\begin{centering}
\includegraphics[scale=0.26]{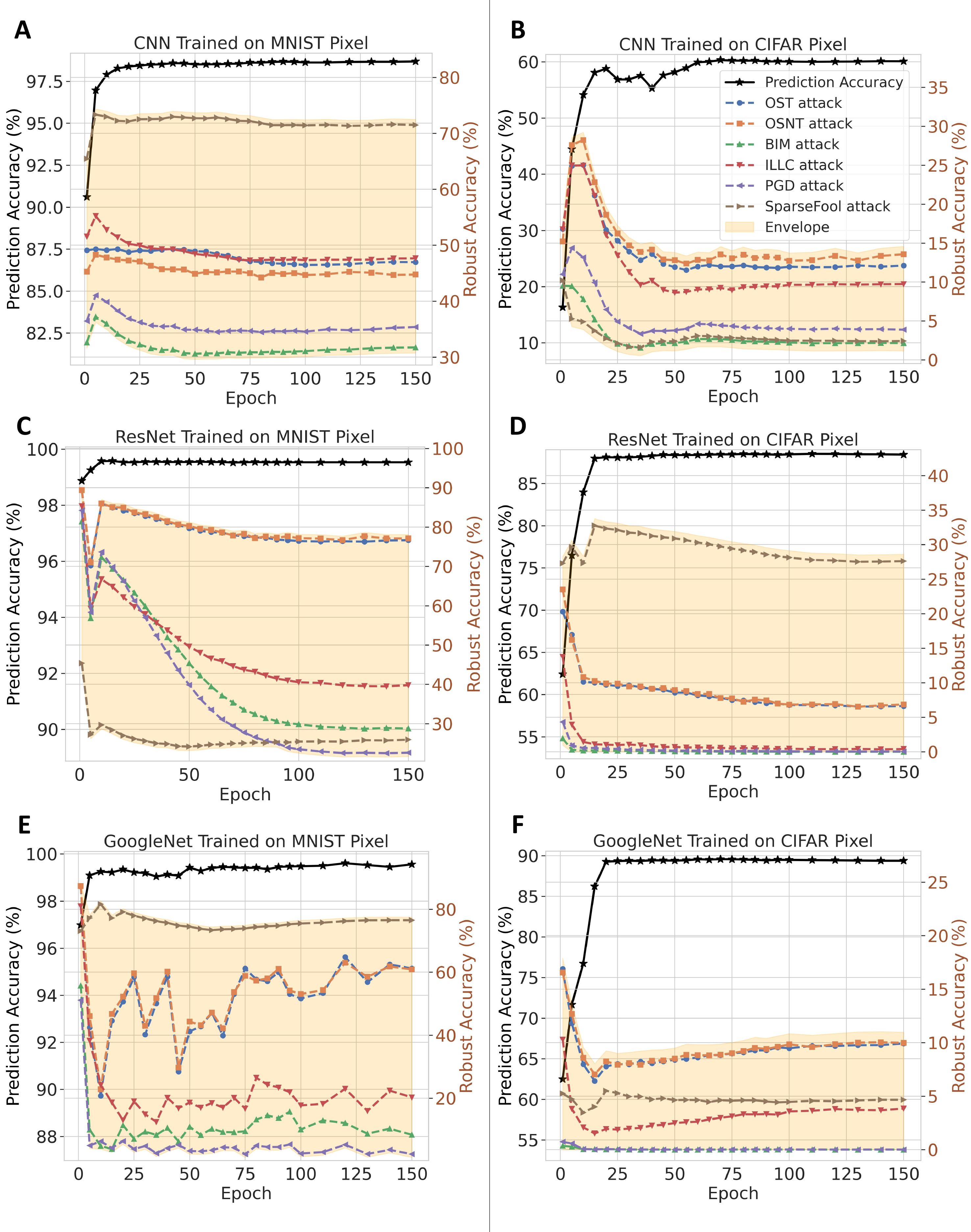}
\par\end{centering}
\caption{Test accuracy and robust accuracy under different attack methods for the MNIST and CIFAR-10 datasets. Networks are trained using a 4-layer CNN, ResNet, and GoogleNet. The black lines represent test accuracy, while the other six colored lines indicate robust accuracy. The attack amplitude for MNIST is 0.1, whereas for CIFAR-10 it is 8/255. \label{fig:various_attack_pixel}}
\end{figure}

To see the trade-off between test and robust accuracies across various attack methods, we train three neural networks separately on the MNIST and Cifar-10 datasets: (1) a 4-layer CNN, (2) ResNet, and (3) GoogleNet. Here, robust accuracy refers to the accuracy when the network input is attacked. Due to the varying performances of different networks and the distinct pixel distributions of the datasets, the test accuracies of these networks versus epochs are shown by the black lines in Fig. 2. For the MNIST dataset all three neural networks achieve test accuracies over 98\%. For the Cifar-10 dataset the simple CNN achieves only around 60\% test accuracy, while ResNet and GoogleNet reaches nearly 90\%.

We then subject all networks to six types of gradient-based attacks: one-step target (OST) attack \cite{goodfellow2015explaining}, one-step non-target (OSNT) attack \cite{goodfellow2015explaining}, basic iterative method (BIM) attack \cite{kurakin2017adversarial}, iterative least-likely (ILLC) class attack \cite{kurakin2017adversarial}, projected gradient descent (PGD) attack \cite{madry2018towards}, and SparseFool attack \cite{8954332}. The other six lines in the figure represent the robust accuracies versus epochs after applying these attacks, and we draw a robust accuracy envelope to show the general variation trend. As observed, the general trend of test accuracy and robust accuracy is opposite: test accuracy increases with epochs, while robust accuracy under attack decreases simultaneously. Additionally, since the MNIST dataset is less complex, the robust accuracy only decreases by around 10\% to 15\% after various attacks. However, for the Cifar-10 dataset, the situation is more severe, with a decrease in robust accuracy of approximately 30\% to 50\%. Furthermore, more complex neural networks trained on more intricate datasets are more susceptible to attacks.

\subsection{\textbf{Phenomena in language model}}

\begin{figure}
\begin{centering}
\includegraphics[scale=0.155]{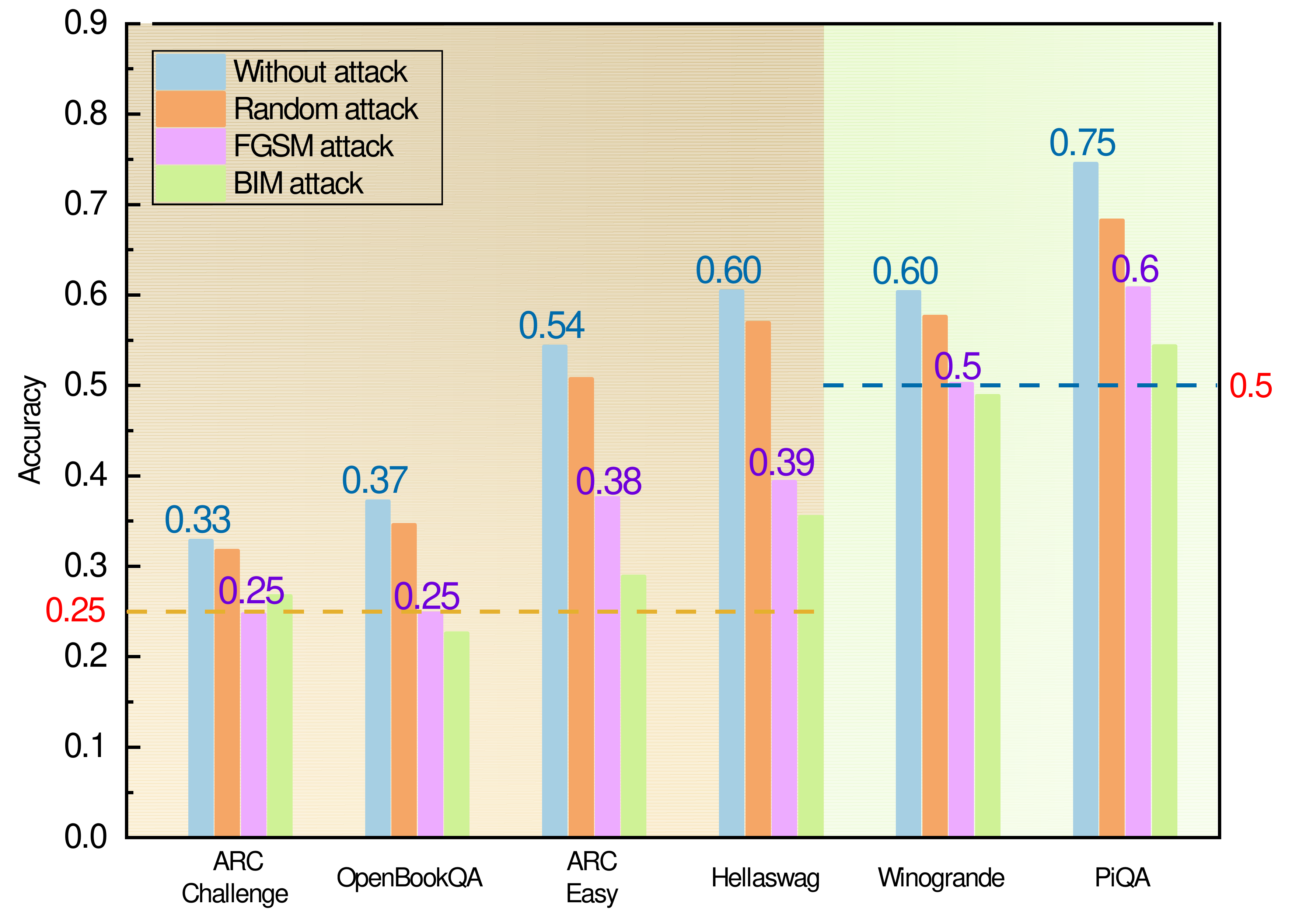}
\par\end{centering}
\caption{Performance of the TinyLlama model on six representative multiple-choice datasets under four distinct conditions. The background is color coded to distinguish dataset types: yellow for four-option datasets and green for two-option datasets. Dashed lines represent random guess accuracy (0.25 for four-option datasets and 0.5 for two-option datasets). Bars indicate the model's evaluation accuracy, with annotations highlighting accuracy without attacks and the FGSM attacks.\label{fig:LLM_attack}}
\end{figure}

To assess the vulnerability of large language models (LLM) under small perturbation attacks, we conducted experiments in open-source LLM TinyLlama \cite{zhang2024tinyllamaopensourcesmalllanguage} by introducing three types of perturbation into its embedding layer output: Random attack, FGSM and BIM. Specifically, FGSM and BIM are gradient-based attacks, whereas Random attack introduces gradient-independent random noise to the embedding vectors. All attacks were applied with a fixed amplitude of 0.01. The random and FGSM attack have the same amplitude and only differ in their directions. Subsequently, we evaluated the model's performance using the Language Model Evaluation Harness \cite{eval-harness} (LM-Evaluation-Harness) on six multiple-choice benchmark datasets: ARC Challenge \cite{clark2018think}, ARC Easy, Hellaswag \cite{zellers2019hellaswag}, OpenBookQA \cite{2018Can}, PiQA \cite{bisk2020piqa}, and Winogrande \cite{Sakaguchi2019AnAW}. 

The experimental results, as shown in Fig. \ref{fig:LLM_attack}, provide insights into the model's behavior under these attack scenarios. We see that all three types of attacks lead to a degradation in the performance of the model. However, gradient-based attacks cause significantly larger performance drops compared to random perturbations. In fact, for certain datasets, such as ARC Challenge, OpenBookQA, and Winograde, the model performance under gradient-based attacks decreases to the random guessing level. For instance, in the ARC Challenge dataset, where each question has four options, the model's accuracy drops to 0.25 under both gradient-based attacks, indicating that it performs no better than random guessing. Similarly, for the Winograde dataset, where each question has two options, the model's accuracy decreases to 0.5, again corresponding to the random guessing baseline.

\section{\textbf{Results: Neural networks exhibit the principle of complementarity similar to that found in quantum mechanics}}

We introduce an uncertainty principle for deep neural networks by drawing an analogy with the Heisenberg uncertainty principle in quantum mechanics. This principle describes an inherent trade-off between a neural network's ability to recognize two conjugate variables: the input features and the corresponding gradients (gradients of the loss function with respect to these features). Based on the uncertainty principle, we demonstrate that the trade-off between the network's prediction accuracy and its robustness to adversarial perturbations is a natural result of the uncertainty relation.

\subsection{\textbf{Uncertainty principle paralleled from quantum mechanics for neural networks}}

We define a 'neural packet' $\psi_Y(X)$ analogous to a quantum wave function, which is the normalized loss function of a network in response to input $X$ and target $Y$,
\begin{equation}
\psi_{Y}(X) = \frac{l(f(X,\theta),Y)}{\sqrt{\beta}},
\end{equation}
where the normalization coefficient \( \beta \) ensures that the integral of the squared neural packet over its input space equals one, $\int \psi_{Y}(X)^{2}dX=1$. 

To obtain the uncertainty relation, we use the Dirac notation in quantum physics in the following contexts. For an input $X=(x_{1},...,x_{i},...,x_{M})$ with $M$ components in the multi-dimensional space, we introduce two operators, named the feature and attack operators, as follows,
\begin{eqnarray}
\hat{x}_{i}\psi_{Y}(X) & = & x_{i}\psi_{Y}(X),\nonumber \\
\hat{p}_{i}\psi_{Y}(X) & = & \frac{\partial}{\partial x_{i}}\psi_{Y}(X).\label{eq:operator_nu}
\end{eqnarray}
Since $\psi_Y(X)$ is square normalized, its square can be seen as a ``probability density'', allowing us to calculate the corresponding mean values of the two operators, 
\begin{eqnarray}
\langle\hat{x}_{i}\rangle & = & \int \psi_{Y}(X)x_{i}\psi_{Y}(X))dX.\\
\langle\hat{p}_{i}\rangle & = & \int \psi_{Y}(X)\frac{\partial}{\partial x_{i}}\psi_{Y}(X)dX.\label{eq:ave_x_nu}
\end{eqnarray}
It can be proved that the standard deviations $\sigma_{a}$ and $\sigma_{b}$
corresponding to two general operators $\hat{A}$ and \textbf{$\hat{B}$} obey the following uncertainty relation (derivation is provided in the supplementary information),
\begin{eqnarray}
\sigma_{a}\sigma_{b} = \langle(\hat{A}-\langle\hat{A}\rangle)^{2}\rangle^{\frac{1}{2}}\langle(\hat{B}-\langle\hat{B}\rangle)^{2}\rangle^{\frac{1}{2}}& \geq & |i\frac{1}{2}\langle[\hat{A},\hat{B}]\rangle|.
\label{eq:sasb}
\end{eqnarray}
Thus, in the context of neural networks, we can replace operators $\hat{A}$
and $\hat{B}$ by $\hat{p}_{i}$ and $\hat{x}_{i}$,
\begin{eqnarray}
\sigma_{{p}_{i}}\sigma_{{x}_{i}} & \geq & |i\frac{1}{2}\langle[\hat{p}_{i},\hat{x}_{i}]\rangle| = \frac{1}{2},\label{eq:uncertainty_relation}
\end{eqnarray}
where we have used the relation
\begin{eqnarray}
[\hat{p}_{i},\hat{x}_{i}]\psi_{Y}(X) & = & [\hat{p}_{i}\hat{x}_{i}-\hat{x}_{i}\hat{p}_{i}]\psi_{Y}(X)\nonumber \\
& = & \frac{\partial}{\partial x_{i}}[x_{i}\psi_{Y}(X)] \nonumber \\
& & -x_{i}\frac{\partial}{\partial x_{i}}\psi_{Y}(X)\nonumber \\
 & = & \psi_{Y}(X).\label{eq:eigencommutator}
\end{eqnarray}

The lower bound in Eq. (\ref{eq:uncertainty_relation}) arises from the normalization of the loss function, which is influenced by both the dataset and the network architecture. Consequently, for a trained neural network, the uncertainty relation is inherently dependent on the specific task (defined by the dataset) and the structure of the network.

\subsection{\textbf{Explanation of the uncertainty principle of neural networks}}

\begin{figure}
\begin{centering}
\includegraphics[scale=0.22]{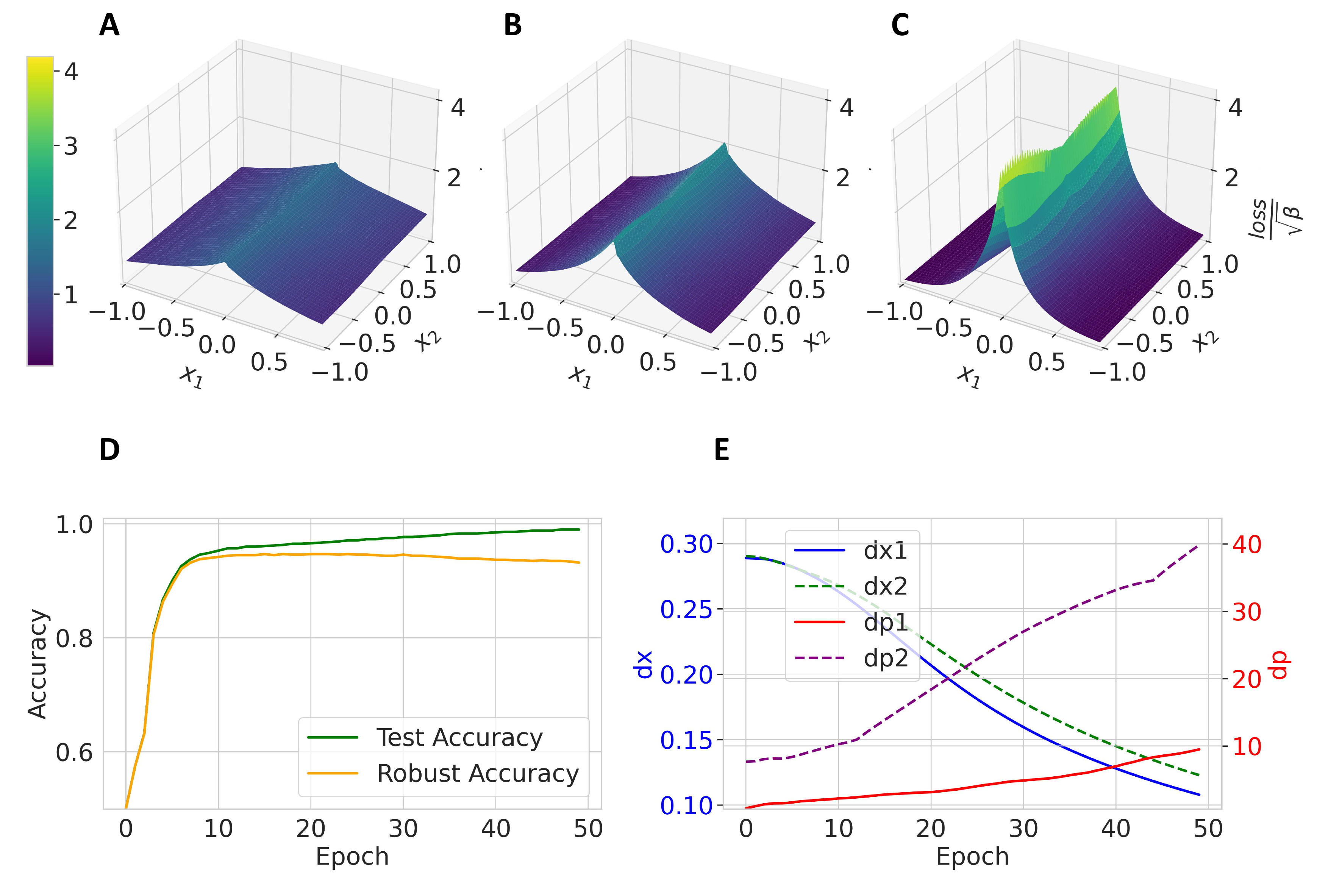}
\par\end{centering}
\caption{Explanation of the uncertainty relation via the binary classification. A, B and C give the loss landscapes at different training epochs 100, 200, and 500. The x and y axes represent the input features $x_1$ and $x_2$, respectively, while the z axis represents the normalized loss value $\psi(x1, x2, Label)=\frac{\text{loss}}{\sqrt{\beta}}$. These plots demonstrate how the loss landscape evolves during training, becoming sharper as the model converges. D shows the test accuracy (green) and robust accuracy under gradient attack (orange) as a function of training epochs. E presents the evolution of $dx$ and $dp$ values, which quantify the test accuracy and sharpness of the loss landscape. For convenience, here $dx$ and $dp$ are evaluated with $x_2=0$. Since the category interface between the two classes is similar along the $x_2$ direction, for convenience, we select a specific value for the loss function, i.e., $\psi(x1, x2=0, Label)$. These plots highlight the trade-off between model accuracy and robustness, as well as the relationship between loss landscape sharpness and adversarial vulnerability. The results are averaged among 5 random seeds for reproducibility. \label{fig: binary_results}}
\end{figure}

\textbf{Intuitive explanation.} We use a binary example to explain the meaning of the quantities $\sigma_{x}$ and $\sigma_{p}$.
A simple neural network is employed to classify data points based on their position relative to the origin. The dataset consists of points uniformly distributed in a 2D space, with labels assigned based on whether the points lie on the positive or negative side of the $x$-axis. The neural network model comprises multiple fully connected layers and is trained over 500 epochs. Various metrics, including loss, accuracy, $\sigma_{x}=dx$ and $\sigma_{p}=dp$, are tracked throughout the training process. To ensure the reliability of our results, the experiment is repeated five times with different random seeds, and the average values are computed.

The results are presented in Fig. \ref{fig: binary_results}. We observe that as test accuracy increases, robust accuracy gradually decreases. The boundary separating the two categories becomes steeper, leading to a decrease in $dx$. Concurrently, as $dx$ decreases, the quantity $dp$ increases, indicating the network's vulnerability. This is because $dp$ measures the sharpness of the boundary, which can also be understood from the Fourier transformation of the gradient:
\begin{equation}
\mathcal{F}\left\{\frac{\partial \psi_{Y}(x)}{\partial x}\right\} = \int_{-\infty}^{\infty} \frac{\partial \psi_{Y}(x)}{\partial x} e^{-i p x} \, dx = i p \hat{\psi}_{Y}(p). \label{eq:fourier_gradient}
\end{equation}
In real scenarios, the inputs often contain many concepts and features at different levels. Therefore, we encounter tremendous sharp boundaries that make the network vulnerable. $dp$, however, provides us with a quantitative measure of these many sharpnesses.

\textbf{Quantum explanation.} There are many concepts that are vital in quantum physics, such as wave packet, quantum state, uncertainty principle, quantum entanglement, state collapse, etc. Analogously, in the context of neural networks, even if it is not designed to mimic a quantum system, it can manifest some quantum aspects, such as the uncertainty principle and state collapse, to help us understand vulnerability phenomena and complementary effects. 

We can treat the images that can be understood by human beings as eigenvalues of a neural network, as defined in Eq. (\ref{eq:operator_nu}). In the surprisingly large space expanded by the network input, there are tremendous images, but only those that mimic the real world and can be understood by human beings are the meaningful ones. In classification tasks, the classifier is fed with the eigenvalues (real images), and the network learns to mimic an operator where the normalized loss functions are the eigenstates  (also called ontological states \cite{hooft2016cellular}), as defined in Eq. (\ref{eq:operator_nu}). Similarly, for generative tasks, the neural network contains two procedures: to understand the inputs (similar to the classification task) and to produce correct outputs that can be understood by human beings. The latter can be seen as an artificial state collapse, hence a measurement. These analogies are essential for deriving the uncertainty relation, thus we have defined and given meanings to these quantities in consistency with Eq. (\ref{eq:operator_nu}). 

The concept of a neural packet is quite similar to the concept of a wave packet. Since a neural packet is also square normalized to be unity, its square has the meaning of probability density. To see this, we again use the binary classification example. For images that have a higher probability of being recognized correctly, their squared loss values should be smaller. Those images near the boundaries are the most likely ones to be incorrectly categorized, hence they have larger square values. It is worth noting that, contrary to the wave packet where larger square values indicate higher probability, the neural packet should be understood reversely in order to be consistent with the uncertainty relation.

The uncertainty relation is a genuine quantum phenomenon, which has a definite meaning in neural networks in terms of the complementarity principle. We can think of two aspects in understanding the images of the same network: one with loss landscape in the feature (input) space and one with loss landscape in the conjugate space. In both landscapes, accurate results should correspond to small dx or dp. Since accurate neural networks indicate smaller dx in feature space but result in larger dp in conjugate space, one cannot expect to obtain high accuracy in both feature and conjugate spaces. The attack method actually creates a scenario where information in feature and conjugate space is forced to be identified simultaneously (Eq. (\ref{eq:fourier_gradient})), leading to a decline in accuracy.

However, it remains an open question whether one needs to treat neural networks as a non-linear version of a quantum system, since even in quantum mechanics there are still studies that consider quantum mechanics as a cellular automaton \cite{hooft2016cellular}. Regardless of these arguments, it is beneficial, at least at the level of mathematics, for us to use the uncertainty relation in Eq. (\ref{eq:uncertainty_relation}) to perceive the complementary aspects of neural network vulnerability and provide different angles with conclusions that cannot be drawn by other theories. For instance, in training language models, if the training data contain complementary features, we may encounter abrupt collapse. In video generation, if the output is a complementary feature of the input, the correct physical laws cannot be learned by a single model. However, at the current stage, we still need more studies before we can conclude that neural networks are worth being mathematically modeled as quantum systems or other quantum effects can be added to the network design for robustness \cite{doi:10.1073/pnas.1324238111}.

\subsection{\textbf{Uncertainty principle essentially works at the feature level}}

\begin{figure}
\begin{centering}
\includegraphics[scale=0.34]{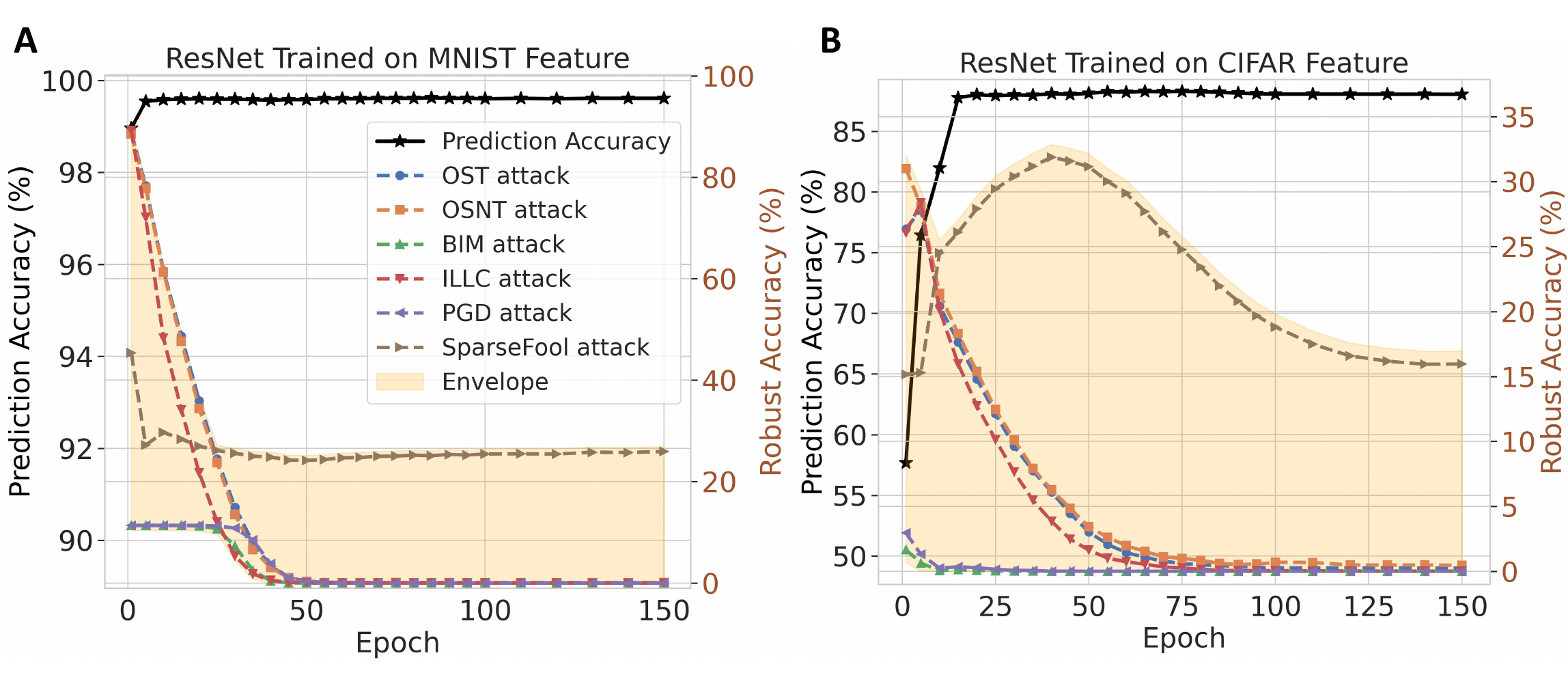}
\par\end{centering}
\caption{Test accuracy and robust accuracy under different attack on feature levels. The ResNet for the MNIST and CIFAR-
10 datasets in Fig. \ref{fig:various_attack_pixel} is divided into two parts: feature extractor and subsequent classifier. All attacks are performed on the outputs (hence, inputs of the subsequent classifier) of the feature extractor. \label{fig:various_attack_feature}}
\end{figure}

It is natural to question the discrepancy between the observed phenomena in Fig. \ref{fig:various_attack_pixel} and the uncertainty relation derived in Eq. (\ref{eq:uncertainty_relation}), given that the pixels in the input images are dependent, while the uncertainty relation applies to an arbitrary dimension. In reality, the uncertainty relation operates at the level of features rather than individual pixels. As previously discussed, the uncertainty principle arises from the sharp boundaries between concepts and features, indicating that complementarity should manifest at the feature level and their corresponding conjugates. This is why we refer to the operator $\hat{x}_{i}$ as a feature operator rather than a pixel operator. Since the features of the input can be considered relatively independent variables, the uncertainty principle is more pronounced at the feature scale.

To illustrate this, we conducted experiments similar to those in Fig. \ref{fig:various_attack_pixel}, but with the attack applied to the feature layers of the neural networks instead of the inputs. Because the specific features of the images cannot be exhaustively enumerated, we divided the neural network into two parts: the initial few layers, which extract features, and the subsequent layers, which perform classification. The attack was primarily executed on the outputs of the initial part (which also serve as the inputs to the latter part). The results are shown in Fig. \ref{fig:various_attack_feature}. Compared to Fig. \ref{fig:various_attack_pixel}, the decline in robust accuracy with training epochs is more pronounced, supporting the underlying uncertainty relation.

\subsection{\textbf{Role of complementarity in network optimization}\label{complementarity}}

\begin{figure}
\begin{centering}
\includegraphics[scale=0.39]{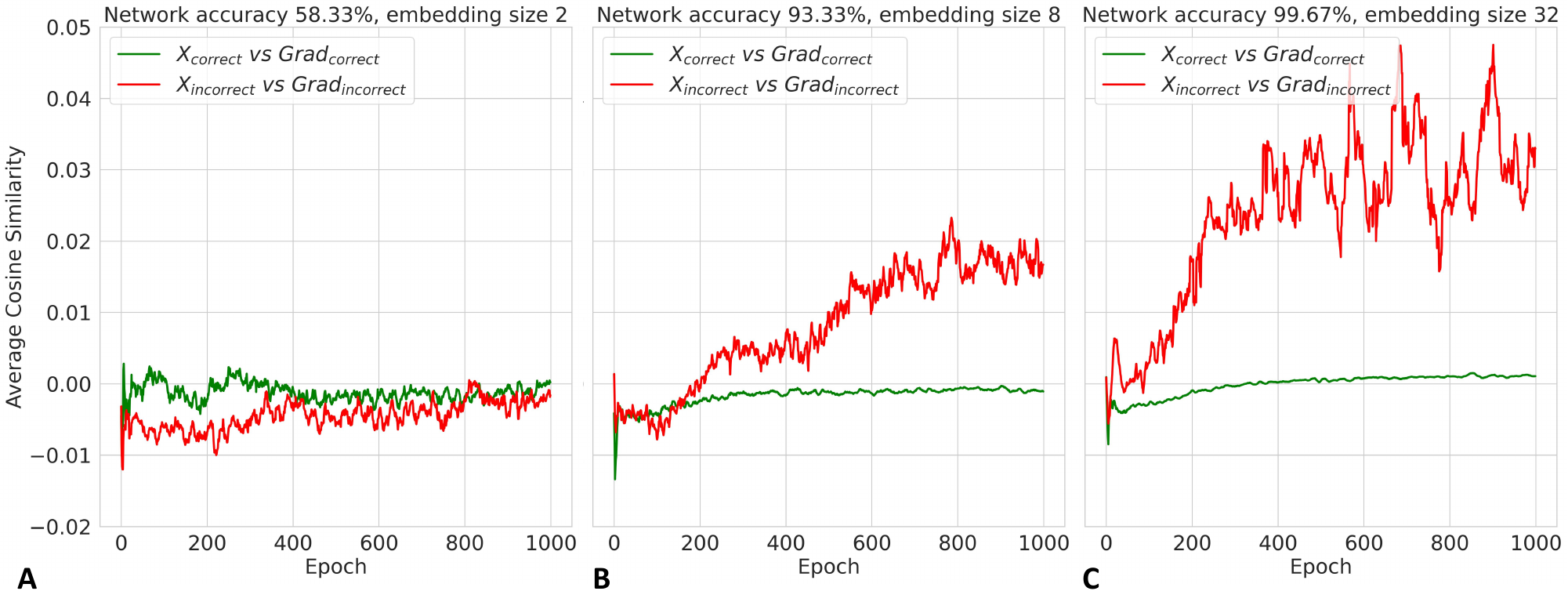}
\par\end{centering}
\caption{Cosine similarities between inputs and gradients. Three types of words (positive, negative, and neutral) are fed to the neural network for sentiment classification. The three neural networks are differ in the embedding size of the words. For each network model, similarities between correct input and the corresponding gradient, as well as the similarities between incorrect input and relevant gradients are calculated at each epoch. The curves are averaged among 10 different random seeds for reproducibility. \label{fig:similarities}}
\end{figure}

The uncertainty principle of neural networks suggests that highly accurate networks tend to reach the lower bound of the uncertainty relation, which is influenced by the specific task and network structure. This principle can be understood through the complementarity principle \cite{Bohr1950}, which addresses the mutually exclusive nature of two conjugate variables. In neural networks, these conjugate variables are the input and the gradient (gradient of the loss function with respect to the input). Consequently, we can anticipate that inputs correctly identified by the neural network should be as irrelevant as possible to the corresponding gradients, i.e., the neural network tends to avoid recognizing two conjugate variables simultaneously. Conversely, inputs that cannot be correctly identified must share more common features with the relevant gradients, so the network fails in recognizing these inputs correctly. This phenomenon should be more prominent in more accurate networks.

To verify this hypothesis, we conducted an experiment involving word classification, categorizing words into positive, negative, and neutral sentiments. The dataset included a mix of English and Chinese words, which were labeled accordingly. These words were encoded and embedded into high-dimensional embedding vectors, and were fed into a neural network model consisting of multiple fully connected layers for classification. The model was trained over 1000 epochs with different embedding sizes to control the model accuracy. We then computed the cosine similarities between the input embeddings and their corresponding gradients for both correct and incorrect inputs, resulting in two types of similarities: correct input vs. correct gradient and incorrect input vs. incorrect gradient. The average cosine similarities for each category were tracked over the training epochs and collected for multiple random seeds to ensure reproducibility. The results, presented in Fig. \ref{fig:similarities}, confirm our anticipation: more accurate neural networks exhibit a more pronounced effect limited by the uncertainty relation, as manifested by the similarity between incorrect inputs and their corresponding gradients. Meanwhile, the similarity between correct inputs and the gradients tends to be zero for all models.

\section{Conclusion}

Our work reveals that neural networks are subject to an uncertainty relation, similar to the uncertainty principle in quantum mechanics, through mathematical proofs and the concept of complementarity. We demonstrate that neural networks exhibit an inherent trade-off between accuracy and robustness, which can be quantified using the uncertainty relation. This trade-off explains why highly accurate neural networks tend to be more vulnerable to adversarial attacks. 

Furthermore, our analysis shows that the vulnerability of neural networks is fundamentally due to the formation of sharp boundaries when learning different class concepts. Additionally, we use the complementarity principle to explain the difficulty in further optimizing highly accurate networks.

Our work suggests  new directions for designing robust neural networks, such as exploring new architectures that can significantly separate concepts, or incorporating methods like ohm’s global quantum potential \cite{doi:10.1073/pnas.1324238111} to improve the network's handling of sharp structures between concepts.

\section{Acknowledgments}
We thank Prof. David Donoho from Stanford University for providing
valuable suggestions on the accuracy robustness of neural networks.

\section{Funding}

The work is partly
supported by the National Natural Science Foundation of China (NSFC) under
the grant number 12105227 and 12405318, and the National Key Research and Development Program of China under Grant No. 2020YFA0709800.

\section{Preprints}
A preprint of this article is published at

https://doi.org/10.48550/arXiv.2205.01493.

\section{Data availability}
All data are available in the main text or the supplementary materials. Additional data related to this paper are available at https://doi.org/10.7910/DVN/SWDL1S

\bibliography{sn-bibliography}% common bib file
%% if required, the content of .bbl file can be included here once bbl is generated
%%\input sn-article.bbl

\end{document}